\documentclass{article} %
\usepackage[final]{corl_2020_modified} %

\usepackage{hyperref}
\usepackage{url}

\usepackage{microtype}
\usepackage{graphicx}
\usepackage{subfigure}
\usepackage{booktabs} %
\usepackage{wrapfig}  %
\usepackage[font={small}]{caption}
\usepackage{algorithm}
\usepackage[noend]{algpseudocode}

\usepackage{enumitem}
\setitemize{noitemsep,topsep=0pt,parsep=0pt,partopsep=0pt}

\newcommand\figcaption{\def\@captype{figure}\caption} 
\newcommand\tabcaption{\def\@captype{table}\caption}

\newcommand{\myparagraph}[1]{\textbf{#1}~} %

\newcommand{\mytimes}{\medmuskip=0mu\times}
\newcommand{\dense}{\thickmuskip=2mu}

\newcommand{\proprio}{o^{\mathrm{pr}}}
\newcommand{\visual}{o^{\mathrm{cam}}}
\newcommand{\topdown}{o^{\mathrm{top}}}
\newcommand{\pose}{o^{pos}}
\newcommand{\boxposes}{o^\mathrm{box}}
\newcommand{\completion}{\beta}
\newcommand{\abstractstate}{S}
\newcommand{\estabstractstate}{S}
\newcommand{\trueabstractstate}{{S}^{*}}
\newcommand{\instruction}{A}
\newcommand{\numactors}{N_{\mathrm{actor}}}
\newcommand{\controltimelimit}{T_{instr}}

\newcommand{\perception}[1]{\mathrm{\textbf{Perception}}}
\newcommand{\planner}[1]{\mathrm{\textbf{Planner}}}
\newcommand{\controller}[1]{\mathrm{\textbf{Controller}}}
\newcommand{\abstractbuffer}{B^{\mathrm{abst}}}
\newcommand{\physicalbuffer}{B^{\mathrm{phy}}}
\newcommand{\obs}{o}

\newcommand{\terminal}{terminal}

\newcommand{\action}{a}
\newcommand{\controlreward}{r}
\newcommand{\abstractstep}{T}

\newcommand{\false}{\mathrm{False}}
\newcommand{\func}[1]{\mathrm{#1}}

\newcommand{\secref}[1]{Section~\ref{#1}}
\newcommand{\eqref}[1]{(\ref{#1})}
\newcommand{\figref}[1]{Fig.~\ref{#1}}

\newcommand{\tabref}[1]{Table~\ref{#1}}

\def\abovestrut#1{\rule[0in]{0in}{#1}\ignorespaces}

\newcommand{\smallgap}{\abovestrut{0.15in}}

\newcommand\tablescaler{0.8}
\title{Beyond Tabula-Rasa: \\a Modular Reinforcement Learning Approach for Physically Embedded 3D Sokoban}

\author{Peter Karkus$^{1,2}$\thanks{Work done at DeepMind.}, Mehdi Mirza$^{2}$, Arthur Guez$^{2}$, Andrew Jaegle$^{2}$, Timothy Lillicrap$^{2}$, \\
\textbf{Lars Buesing$^{2}$, Nicolas Heess$^{2}$, Theophane Weber$^{2}$} \\
$^1$National Unviersity of Singapore, $^2$DeepMind\\
	\texttt{karkus@comp.nus.edu.sg} %
}

\let\OLDthebibliography\thebibliography
\renewcommand\thebibliography[1]{
  \OLDthebibliography{#1}
  \setlength{\parskip}{4pt}
  \setlength{\itemsep}{0pt plus 0.3ex}
}

\begin{document}

\maketitle

\begin{abstract}
Intelligent robots need to achieve abstract objectives using concrete, spatiotemporally complex sensory information and motor control. \emph{Tabula rasa} deep reinforcement learning (RL) has tackled demanding tasks in terms of either visual, abstract, or physical reasoning, but solving these jointly remains a formidable challenge.
One recent, unsolved benchmark task that integrates these challenges is Mujoban, where a robot needs to arrange 3D warehouses generated from 2D Sokoban puzzles.
We explore whether integrated tasks like Mujoban can be solved by composing RL modules together in a \emph{sense-plan-act} hierarchy, where modules have well-defined roles similarly to classic robot architectures. Unlike classic architectures that are typically model-based, we use only model-free modules trained with RL or supervised learning. We find that our modular RL approach dramatically outperforms the state-of-the-art monolithic RL agent on Mujoban. Further, learned modules can be reused when, e.g., using a different robot platform to solve the same task. 
Together our results give strong evidence for the importance of research into modular RL designs. Project website: {\footnotesize \url{https://sites.google.com/view/modular-rl/}}
\end{abstract}

\keywords{Hierarchical reinforcement learning, planning, partial observability}  %

\section{Introduction}
\label{introduction}

\begin{wrapfigure}{r}{.40\textwidth}
\vspace*{-12pt}
\centering
\includegraphics[width=.96\linewidth]{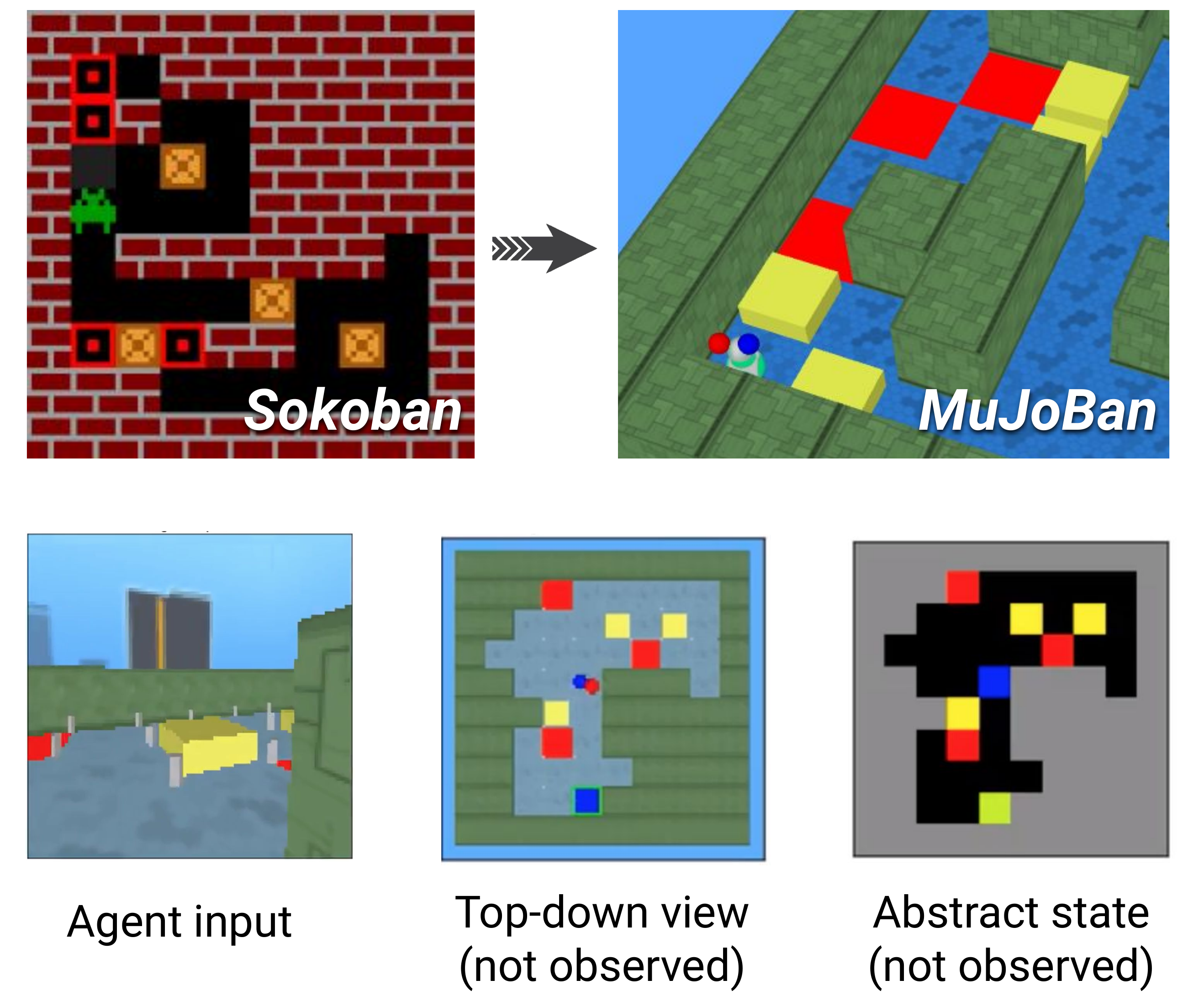}%
\caption{\small Mujoban is a challenging embodied  task that embeds Sokoban puzzles in a physical 3D simulator~\citep{mujoban}.
The robot receives partial, first-person observations and has to push boxes (yellow) onto targets (red).\vspace{-8pt}
}
\label{fig:mujoban}
\end{wrapfigure}

Deep RL has successfully tackled tasks involving complex motor control~\cite{heess2017emergence,gupta2019relay,andrychowicz2020learning}, long-horizon abstract reasoning~\cite{silver2016mastering,berner2019dota,vinyals2019alphastar}, as well as partial and visual observations~\cite{levine2016end, sadeghi2017cad, gregor2019shaping, wijmans2019dd, ma2020discriminative}. %
But these challenges have largely been tackled in isolation. Moving forward, an important question is how to best utilize deep RL techniques for building real world intelligent robot systems -- such as, for instance, a robot that cooks and cleans up. All the above challenges now need to be addressed \emph{jointly} in a single system. Several integrated benchmark domains have recently been proposed to facilitate research in this direction~\cite{lee2019ikea, mujoban}.
Unfortunately, deep RL results in these domains are currently very poor. For example, in the Mujoban domain~\cite{mujoban} a robot needs to solve physically embedded Sokoban puzzles by pushing boxes in a maze given partial, first-person visual observations~(\figref{fig:mujoban}). State-of-the-art RL can solve less then 10\% of the trials even after weeks of training on TPUs and with access to privileged information such as a top-down view~(see~\cite{mujoban}). %

This paper investigates whether difficult embodied tasks can be approached by \emph{composing} RL modules together similarly to classic robot architectures. Specifically, we focus on the Mujoban domain and explore a modular RL architecture with a sense-plan-act hierarchy~(see \citealt{arkin1998behavior}, Ch. 4), where separate modules perform perception (``sense''), abstract reasoning (``plan''), and low-level motor control (``act'').  
In contrast to most hierarchical RL approaches~\citep{bacon2017option} we carefully prescribe the role of each module in the hierarchy,
aiming to build in structure specific to embodied tasks, as well as knowledge about useful abstractions. %
This modular design allows us to choose effective network structures and training regimes, and it also allows us to reuse trained modules, e.g., for different tasks or on different robot platforms. 
Modern robot architectures often combine model-based and learned components~\citep{andrychowicz2020learning}, but to the best of our knowledge, exclusively model-free RL modules in a sense-plan-act hierarchy have not been studied at the scale of integrated tasks like Mujoban.

\begin{figure}[tb]
\centering
\includegraphics[width=0.90\textwidth]{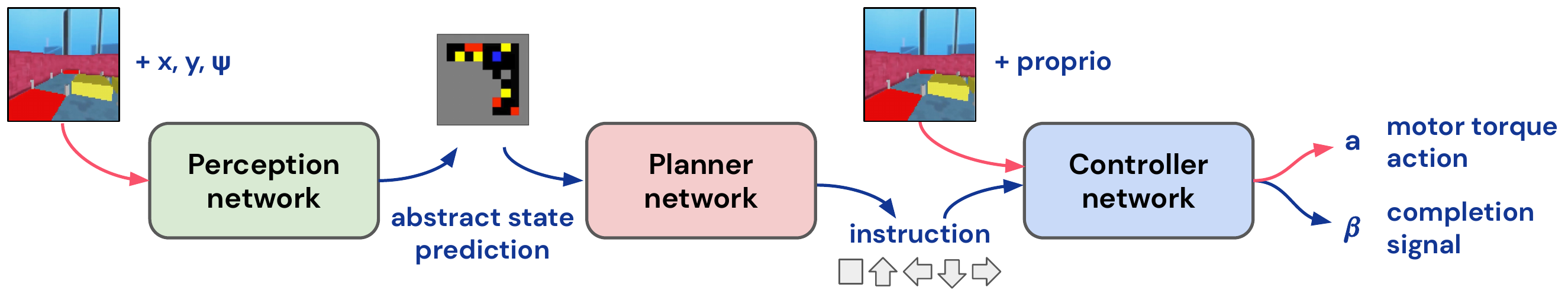}
\caption{\small The proposed modular RL architecture composes separate modules for perception, planning, and control (each trained using model-free RL) to produce an overall policy. 
}\label{fig:modules}
\end{figure}

Our modular RL architecture is shown in \figref{fig:modules}. The architecture consists of three modules, which are trained to optimize different objectives: a controller is trained for goal-oriented locomotion and pushing; a planner  is trained for high-level abstract reasoning; and a perception network is trained to infer an abstract 2D state from sequences of first-person RGB inputs. We do not impose modular structure on the abstract reasoning problem itself, but learn it only from physical interactions, for which we introduce a simple time-abstracted RL algorithm based on MPO~\citep{abdolmaleki2018maximum}.
Modular RL improves the state-of-the-art in Mujoban by a large margin (from 9.4\% to 78.7\% success when using only first-person input). We also show that our learned abstract reasoning module transfers when the 2-DoF ball body is replaced with a 8-DoF ant robot. Our results reaffirm the benefit of prior knowledge in the form of modular structural biases and suggest the importance of research in this direction. %

The specific contributions of this paper are as follows. We empirically demonstrate 1) that model-free RL modules can be composed in a sense-plan-act hierarchy and achieve state-of-the-art in Mujoban, an unsolved simulation domain that jointly exhibits important challenges of robot intelligence (partial visual observations, long-horizon planning, %
motor control); 2) that learned model-free RL modules for planning can be reused on different robot platforms. 
3) We introduce a modular RL algorithm that trains hierarchical policies with  time-abstraction.

\section{Related work}
\label{sec:related_work}

The question of how to incorporate domain knowledge and structure into model-free deep RL has been the focus of much recent research. Domain knowledge may appear in various forms, including reward shaping~\cite{ng1999policy, popov2017data}, training curricula~\citep{bengio2009curriculum, florensa2017reverse}, auxiliary tasks~\citep{jaderberg2016reinforcement}, state representation~\citep{oord2018representation}, algorithmic structure~\citep{tamar2016value, farquhar2017treeqn, ma2020discriminative} and memory~\citep{parisotto2018neural, fortunato2019generalization}. In this work we incorporate structure in the form of modularity, leveraging knowledge about embodied reasoning and abstractions in the target domain.

Hierarchical RL (HRL) has been the subject of research for many decades~\citep{dayan1993feudal, sutton1999between, gregor2016variational,heess2016learning, vezhnevets2017feudal, bacon2017option, merel2019hierarchical, merel2019reusable}.
Our modular RL approach is a form of hierarchical RL: we use a hierarchy of RL policies acting at different levels of abstractions and time scales. Importantly, while HRL typically aims to discover useful abstractions from environment interactions, we aim to utilize knowledge about task abstractions as a form of inductive bias, and thus define the role of each module and prescribe fixed interfaces. Learning abstractions provide more flexibility, which naturally comes at the cost of (often prohibitively) lower data efficiency. The wide-spread use of the sense-plan-act architecture in real-world robot system suggests that imposing such hierarchy might be a viable form of inductive bias for a large class of embodied problems.

Modern robot learning systems often combine hand-crafted and/or model-based modules with learned model-free modules. Recent examples include \citet{andrychowicz2020learning} on solving Rubik's cubes with a fixed abstract planner and a learned RL controller; and \citet{chaplot2020learning} on navigation in indoor spaces using a fixed path planner with learned mapping and control modules. In contrast, our modular RL approach uses model-free \emph{learned} modules exclusively, and trains them with reward or supervised signals. Importantly, we learn abstract planning exclusively from physical interactions with the environment, without accessing a model or simulator of the abstract problem. 

Finally, the Differentiable Algorithm Network (DAN) of~\citet{karkus2019differentiable} combines a sense-plan-act hierarchical structure with end-to-end learning. The DAN composes differentiable structures for perception (vision + filtering), planning, and control, and trains them jointly for partially observed map-based navigation tasks. 
The use of modules with well-defined roles in a hierarchy is similar in our work, however, our modules do not encode task-specific algorithmic structure, and our modules are trained with rewards and supervised signals instead of end-to-end expert demonstrations.

We provide a more extensive discussion on modular structures and RL in Appendix~\ref{sec:app_related_work}. %

\section{Mujoban}
\label{sec:mujoban}

We are interested in integrated embodied tasks that jointly exhibit  challenges of visual perception, abstract reasoning and motor control. We choose Mujoban~\citep{mujoban}, a recently proposed benchmark task that embeds Sokoban puzzles~(\citep{junghanns1997sokoban, racaniere2017imagination, guez2019investigation}) in the Mujoco simulator \citep{todorov2012mujoco}. 

\begin{figure}[tb]
\centering
\includegraphics[width=\textwidth]{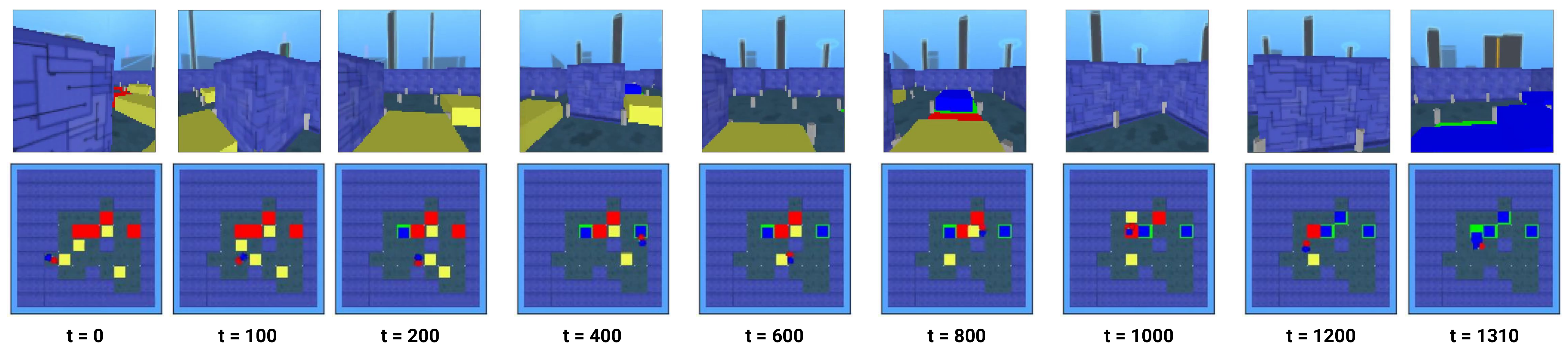}
\caption{\small Modular RL trajectory in Mujoban. We show first-person inputs (top row) and unobserved top-down views (bottom row). Viewed from above, the robot appears as two red and blue discs. Boxes are yellow, target pads are red. When a box is pushed on a target its color changes to blue and the target pad turns green. %
}\label{fig:trajectory}
\end{figure}

Mujoban generates 3D mazes from 2D Sokoban puzzles with randomized visual appearance. A robot navigates the maze and pushes boxes onto target areas with its body. The default robot is a 2-DoF ball, but we also experiment with a 8-DoF ant. We use the harder "peg" version of Mujoban, where small pegs are inserted at grid points to enforce Sokoban rules (boxes cannot be pushed diagonally and cannot be recovered when next to a wall or another box). The robot receives partial observations: first-person camera images $\visual$, proprioceptive signals $\proprio$ (including touch, position, velocity, and acceleration sensors), %
and global pose $\dense \pose=(x, y, \psi)$. Rewards are +10 for solving the puzzle, and +1/-1 for pushes a box on/off a target area.
The challenges of this task are three-fold:
\begin{itemize}
    \item \textbf{Motor control.} The agent needs to learn locomotion (``crawling'' in case of the ant) to reach targets in the maze. It must also push and carefully align boxes, such that they do not get stuck among grid pegs when pushed from another direction. We found this manipulation problem to be challenging on its own.
    \item \textbf{Abstract reasoning / planning.} Sokoban puzzles, by design, require long-horizon planning. Boxes can only be pushed but not pulled, so many moves are irreversible, making it important to plan ahead. 
    \item \textbf{Perception.} Despite the simplistic visual appearance of Mujoban, perception involves non-trivial visual mapping: to infer the underlying 2D Sokoban state, information needs to be integrated from sequences of RGB images showing only parts of the 3D maze.
\end{itemize}
Importantly, different challenges compound: e.g., a poor box alignment may prevent executing a high-level strategy, and imperfect perception may lead to invalid plans with wrong irreversible moves.

In contrast to \cite{mujoban}, and motivated by real-world applications, we target a more realistic partially observable setting of Mujoban. Most importantly, we only provide first-person observations as opposed to first-person and top-down inputs in \cite{mujoban}. %
As is common in robot learning we utilize the simulator more explicitly for training. That is, we access the full physical state to compute rewards, and abstract states $\trueabstractstate$ which serve as supervisory signals. $\trueabstractstate$  captures the underlying Sokoban state in $\dense N\mytimes N\mytimes 4$ binary image (the same $\trueabstractstate$ is input to some agents in~\cite{mujoban}). Finally, we assume the robot pose can be observed for the purpose of mapping, but importantly. we do not observe the environment layout and box poses directly at test time.

\section{Modular Reinforcement Learning}
\label{sec:modular_rl}

\subsection{Overview}
We propose a modular RL approach for Mujoban that decouples layers of embodied reasoning into distinct RL modules: a perception module for spatial reasoning, a planner for abstract reasoning,  and controller for motor control. We prescribe different objectives to each module, leveraging domain knowledge both in terms of generic properties of embodiment as well as task-specific abstractions.

Our modular RL architecture is shown in \figref{fig:modules}. 
The perception module integrates first-person visual information $\obs$ into an abstract 2D state $\estabstractstate$, similar to a 2D map. The planner\footnote{The term \emph{planner} refers to the role of the model-free RL policy, not a model-based planning algorithm.} takes in the abstract state and outputs an instruction $\instruction$.  We design instructions (move north, east, south, west, or stay) knowing that moving and pushing boxes between cells is sufficient to succeed in the task.
Finally, the controller is a goal-oriented visuomotor policy for locomotion and object manipulation (pushing boxes). It takes in observations and an instruction and outputs motor torque actions $\action$. We also predict if the last instruction was completed ($\completion$) and use it for time-abstraction in the planner. %
Modules are trained in different regimes for different objectives, but they rely on each other to collect meaningful experience. Pseudo-code for training modules is given in Algorithm~\ref{alg:modular_rl}. %

Next we introduce each module and the training algorithm. Additional details are in Appendix~\ref{sec:app_module_details}. %

\subsection{Controller module}
The controller is a goal-oriented RL policy for locomotion and object manipulation. The controller also predicts if the last instruction was completed, similarly to the option termination of HRL~\citep{sutton1999between, barreto2019option}. We use separate networks for the control policy and the completion predictor. %

The control policy takes an instruction $\instruction_t$, visual observation $\visual_t$, proprioceptive observations $\proprio_t$, and the (predicted) completion signal $\completion_t$, and outputs continuous motor torque actions $\action_t$. Instructions are one-hot encodings of discrete high-level moves. We use a ResNet~\citep{he2016deep} connected to an LSTM~\citep{hochreiter1997long} and train with the actor-critic MPO algorithm~\cite{abdolmaleki2018maximum}.
We use an asymmetric setup where the critic receives privileged side information during training ($\pose_t$ and box poses $\boxposes_t$).
This training setup is common for robot learning~\cite{pinto2017asymmetric, andrychowicz2020learning}.

We define the following rewards for the controller: +4 if the last instruction is completed and a small negative reward if the robot moves backwards. If the instruction is not completed within $\dense \controltimelimit=120$ steps the agent receives a reward of -5 and the episode terminates. An instruction is defined to be completed only if the robot is near the center of the grid cell for the instructed move, and if pushing a box the box is near the center of the adjacent cell. 

The controller module also outputs a completion signal $\completion_t$, a prediction whether the last instruction was completed. We train a supervised sequential binary classifier, a ResNet-LSTM network independent from the policy network. We obtain training labels given the control rewards defined above. During evaluation a positive completion signal is added if there has been no completion predicted for $\dense \controltimelimit=120$ steps. 

\subsection{Planner module}
The planner is a model-free RL policy responsible for solving the puzzle that underlies the physical environment. Inputs to the planner are abstract state estimates, $\estabstractstate_t$, i.e., $10\mytimes10\mytimes4$ images capturing the 2D maze maze (see~\figref{fig:mujoban} for an example). Outputs are discrete instructions, $\instruction_t$, corresponding to moves between cells of the maze.  The policy acts in abstract time, that is, in each step of the planner the robot may interact with the environment for a variable number of time steps until the instruction is (predicted to be) completed.

We use a policy network similar to the repeated ConvLSTM architecture of \citet{guez2019investigation}. The network is built of generic components, but it has strong structural bias for 2D planning. We use a feed-forward variant of the architecture with an additional critic head that outputs Q-value estimates for each discrete instruction.
The planner policy is trained with RL, but from the planner's perspective it interacts with an abstract version of the environment (through the learned controller) that resembles the symbolic Sokoban domain. Rewards for the planner are task rewards accumulated during the execution of the instruction. The training algorithm is given in Algorithm~\ref{alg:modular_rl} and it is further discussed in
\secref{sec:training}. 

\subsection{Perception module}

\begin{wrapfigure}{r}{.55\textwidth}%
\vspace*{-18pt}
\centering
\includegraphics[width=.52\textwidth]{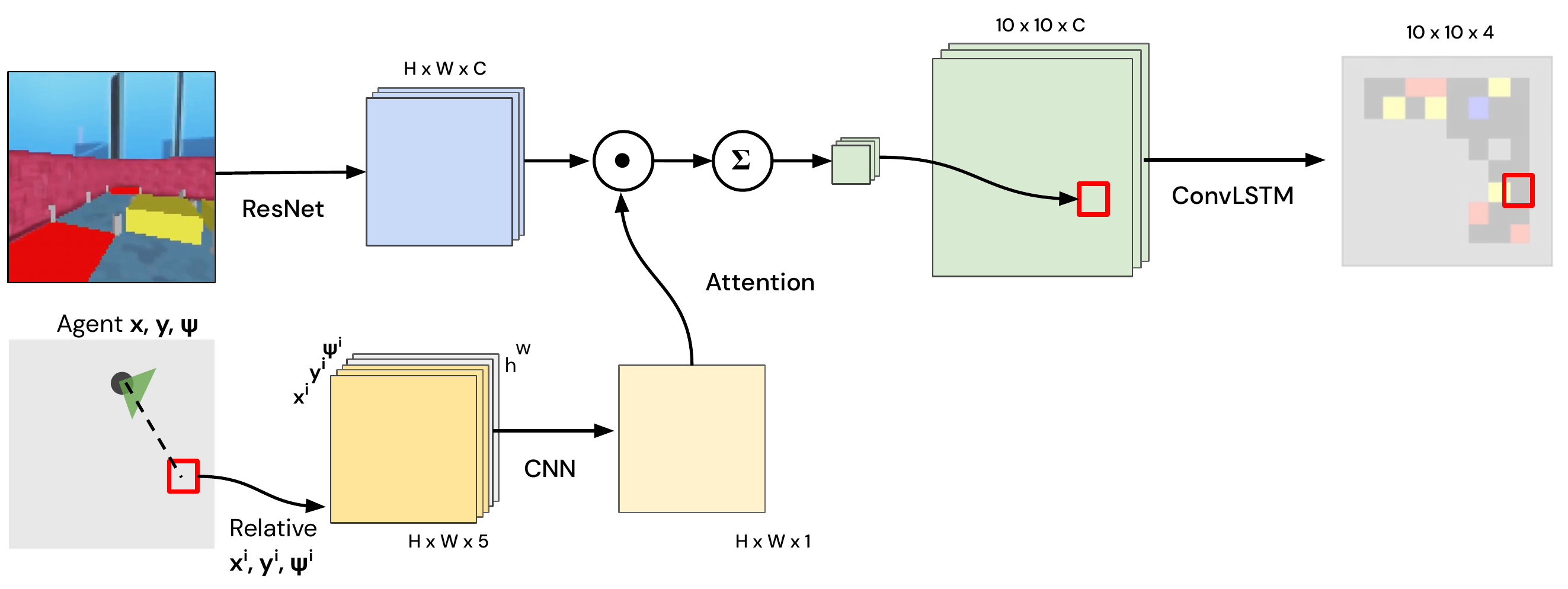}%
\caption{\small Geometry-aware perception network.
}\label{fig:mapping}
\end{wrapfigure}
The perception module sequentially predicts the abstract state  $\estabstractstate_t$ from first-person visual observations $\visual_t$ and the agent poses $\pose_t$. The task is similar to mapping, where $\estabstractstate_t$ is equivalent to a 2D map of the 3D scene. %
We design a geometry-aware network architecture that builds in strong inductive bias for 2D mapping from first-person observations. The network is a ConvLSTM~\cite{xingjian2015convolutional} with a novel spatial attention mechanism shown in \figref{fig:mapping}. The attention mechanism was inspired by~\citep{karkus2020differentiable} but it is adapted to our 2D mapping task. Intuitively, the network works by attending to different elements of the visual input for each possible location of the 2D map ($\estabstractstate_t$). Spatial attention produces features arranged in a 2D grid. The features are fed to a ConvLSTM, forming an implicit belief over abstract states. The LSTM output is fed to a classifier with 5 labels (None,Wall,Box,Target,Box-on-target), which predicts each spatial location of the abstract state separately.
We train the network with independent cross-entropy losses, where labels are abstract state observations  $\trueabstractstate_t$ from the simulator.
The geometry-aware perception network currently relies on observing the agent's pose, %
but it could be also predicted from visual and proprioceptive observations in the future. The network architecture could be also used for 2D mapping tasks beyond Mujoban in future work.

\subsection{Training algorithm}
\label{sec:training}

\begin{table}[h]
\centering
\scalebox{\tablescaler}{
\begin{minipage}{0.95\linewidth}
\begin{algorithm}[H]   %
\caption{Modular RL pseudo code for training modules with time abstraction}
\begin{algorithmic}[1]
\State Initialize $\perception{}, \planner{}, \controller{}$ and replay buffers $\abstractbuffer, \physicalbuffer$
\While{not converged}
    \State $\obs_t  \leftarrow  \func{Reset}()$     \Comment{reset environment}
    \While{not $\terminal$}                                                                \Comment{repeat until episode terminates}  
        \State $\estabstractstate \leftarrow \perception{\theta}(\obs_t)$  \Comment{estimate abstract state}%
        \State $\instruction \leftarrow \planner{\theta}(\estabstractstate)$   \Comment{sample instruction from planner}
        \smallgap
        \State \emph{// execute instruction through multiple environment steps} %
        \State $R \leftarrow 0; \completion_t \leftarrow \false; t \leftarrow 0$  %
        \While{not $\completion_t$ and $t \leq \controltimelimit$ and not $\terminal$}
            \State $\action_t, \completion_t \leftarrow \controller{\theta}(\obs_t, \instruction_\abstractstep)$   %
            \State $\obs_{t+1}, r_t, \terminal \leftarrow$ $\func{Execute}(\action_t$)   \Comment{execute action} %
            \State $R \leftarrow R + r_t$\label{line:abstract_reward}   \Comment{accumulate task reward}
            \State $\func{AddToReplay}(\physicalbuffer, (\obs_t, \action_t, \controlreward_t, \obs_{t+1}, \terminal)$)   %
        \EndWhile
        
        \State $\terminal \leftarrow \terminal$ or $t > \controltimelimit$\label{line:early_termination}  \Comment{early termination}
        \State $\func{AddToReplay}(\abstractbuffer, (\estabstractstate, \instruction, R, \perception{\theta}(\obs_{t+1}), \terminal$)   %
    \EndWhile
    \State \emph{// update modules}
    \State $\func{UpdateRL}(\controller{\theta}, \physicalbuffer)$ \label{line:rl1}  \Comment{uses MPO~\cite{abdolmaleki2018maximum}}
    \State $\func{UpdateRL}(\planner{\theta}, \abstractbuffer)$  \label{line:rl2} \Comment{time-abstracted update, uses MPO~\cite{abdolmaleki2018maximum}}
    \State $\func{UpdateSupervised}(\perception{\theta}, \physicalbuffer)$ \label{line:sl1}     %
\EndWhile %
\end{algorithmic}
\label{alg:modular_rl}
\end{algorithm}
\end{minipage}
}
\end{table}

Algorithm~\ref{alg:modular_rl} provides a pseudo-code for training modules in our architecture. The algorithm addresses the general problem of learning abstract high-level reasoning from physical interactions by collecting time-abstracted experience with a low-level controller that predicts its completion~($\completion$). We define rewards for an abstract step as the sum of environment rewards collected during the execution of the instruction (line \ref{line:abstract_reward}). Failure to complete an instruction within a time limit is treated as an early-termination of the episode (line \ref{line:early_termination}). The algorithm combines RL and supervised updates (lines \ref{line:rl1}, \ref{line:rl2} and \ref{line:sl1}), and thus allows for a pragmatic approach where supervising learning is used where a supervised signal is easy to derive, and RL is used everywhere else.

We omitted a number of task-specific implementation details from Algorithm~\ref{alg:modular_rl} for clarity. We include side information ($\boxposes$, $\trueabstractstate$) in $\obs$ during training which is used to compute controller rewards and supervised losses. We use distributed RL where experience collection and module updates are separated into different nodes. Further, while joint training would be possible, we only train one module at a time in a bottom-up sequence (controller, planner, perception), where untrained modules are replaced with ground-truth or random inputs. Further details are in Appendix~\ref{sec:app_training}. 

All models are implemented in Tensorflow~\citep{tensorflow2015-whitepaper} and trained with the Adam optimizer~\citep{kingma2014adam}.  We use one Nvidia Tesla V100 GPU for learner nodes and CPU only for actor nodes. We train models until near convergence, which took up to 7, 21 and 5 days for modular RL modules, fully-observable baselines and partially-observable baselines, respectively.

\begin{table}[tb]
\centering
\scalebox{\tablescaler}{\begin{tabular}{lccc|cccc}
\toprule
 & \multicolumn{3}{l|}{\textbf{}} &  \textbf{Rewards} & \multicolumn{3}{c}{\textbf{Success rates}} \\
 & \textbf{Percep.} & \textbf{Planner} & \textbf{Contr.} &  \textbf{hard} & \textbf{easy} & \textbf{med.} & \textbf{hard} \\
\midrule  %
Modular RL & - & \emph{oracle} &  learned    & 13.3 \small{(0.1)} & 100\% & 97.9\%& \textbf{94.3\%} \\
Modular RL & \emph{true} & learned  &  learned  & 11.8 \small{(0.2)} & 100\% & 90.0\% & \textbf{81.8\%}  \\
Modular RL & learned & learned &  learned & 11.4 \small{(0.2)} & 90.2\% & 78.7\% & \textbf{78.7\%}  \\
\bottomrule
\end{tabular}
}
\tabcaption{\small Modular RL results with different set of modules.} %
\label{tab:modular_rl_configs}
\end{table}

\section{Results}
We present results for modular RL (\secref{sec:modular_results}) followed by comparison results with alternative learning methods (\secref{sec:comparison_results}) and module transfer results for the Mujoco ant (\secref{sec:ant_results}). %

\subsection{Modular RL results}
\label{sec:modular_results}

\tabref{tab:modular_rl_configs} evaluates trained modules in different configurations: controller only with instructions from an oracle (first row); 
planner and controller with perfect abstract state input (second row); using all learned modules (third row). We report success rates in 512 random episodes for different difficulty levels: easy ($5\mytimes5$ mazes with 1 box), medium ($8\mytimes8$ mazes with 3 boxes), and hard ($10\mytimes10$ mazes with 4 boxes). An episode is successful if the puzzle is solved within 4800 steps (240s). All modules are trained using the default hard level only. Rewards (with standard deviations) are reported for the default difficulty (hard). 
We observe strong performance, $78.7\%$ success in the final setting using all modules (last row). The performance gap without perception is relatively small ($78.7\%$ to $81.8\%$), while the performance gap with and without planning is larger ($81.8\%$ to $94.3\%$).

We report learning curves for each module in \figref{fig:learning_curves}. We plot the relevant objective metric (rewards or accuracy) against environment steps sampled from the replay buffer. %
We selected results for the best hyper-parameter and random seed (out of 2). We continued training for a large number of steps, but the rate of improvements indicate that good performance can be achieved with less training as well.

\figref{fig:trajectory} shows a successful trajectory. Videos are available at {\footnotesize \url{https://sites.google.com/view/modular-rl/}}.
We observe that generally the modular agent quickly builds good abstract representations, moves around and pushes boxes, and solves many levels that are hard (even for a human player with top-down view). 
Failures can be attributed to imperfect abstract states (leading to poor irreversible moves); the planner failing to solve the underlying Sokoban puzzle; and the controller failing an instruction, typically when grid pegs block the motion of a previously poorly aligned box.

Modular RL results are substantially stronger than the best known RL agents for Mujoban (\tabref{tab:sota}). Note that success rates are not directly comparable because these agents receive additional inputs (top-down camera, abstract state, and expert instructions) and are evaluated with a different timeout (45s instead of 240s). The rows of \tabref{tab:sota} correspond to ``Vanilla Agent'', ``Random Planner'' and ``Expert Planner'' of "with grid pegs" results of \citet{mujoban}; and easy, medium and hard levels correspond to level 1, 4, and 5, respectively.

\subsection{Comparison results}
\label{sec:comparison_results}

\myparagraph{Monolithic RL.} 
We first compare modular RL with monolithic RL methods in \tabref{tab:monolithic_comparisons}. We use standard ResNet-LSTM policy networks trained with MPO and the default task rewards. We use the same partially observable setting as for modular RL and the same asymmetric actor-critic setup where the critic receives privileged side information. In different rows of \tabref{tab:monolithic_comparisons} we try a symmetric actor-critic with a shared torso that does not use side information (+shared); training with a curriculum (+curr), or training only on easy levels (+easy-only).
We searched over network architectures of 3 significantly different sizes, learning rates, and MPO parameters. We report only the best performing combinations. Results indicate that monolithic RL can learn almost none but the easiest levels. Further, without modularity, asymmetric actor-critic cannot benefit from  privileged observations; and curriculum helps on easy levels but not on harder ones.

\paragraph{Structured and hierarchical RL}
Prior work proposed various inductive structures for RL that could be useful in this domain, e.g., for mapping~\citep{parisotto2018neural, chaplot2020learning, gupta2017cognitive}, or for state estimation from partial observations~\citep{igl2018deep,ma2020discriminative}. Instead of direct comparisons we run experiments with additional top-down observations that \emph{remove} partial observability and the need for mapping, thus results here can be expected to upper-bound RL approaches with specific structure for mapping or filtering.  For these experiments we follow \citet{mujoban} and use the same network structure, V-trace algorithm, and curriculum. While these choices are orthogonal, we did not try all combinations due to the substantial required computation time.

\tabref{tab:topdown_results} reports results in the simplified top-down setting of Mujoban. %
Monolithic RL can only solve 27.9\% of the hard episodes (first row), which suggests that without top-down input, RL with specific structure for mapping or filtering would not be effective. Next we further simplify the task (second row). We add direct observations of the abstract state and structured exploration, where extra pseudo rewards are given for reaching an adjacent abstract state, similarly to the ``Random planner'' in~\cite{mujoban}. This structured agent performs significantly better, but still worse than our modular approach (77.0\% with additional top-down input vs. 81.8\% for modular RL). %
Interestingly, this agent learned to exploit shortcuts in Mujoban, such as recovering boxes pushed against a wall and push multiple boxes together. %
This suggests that in a stricter version of the task the benefit of the modular approach would be even larger.

\begin{table}[tb]
\begin{minipage}[]{0.65\textwidth}
\centering
\scalebox{\tablescaler}{\begin{tabular}{lcccc}
\toprule
 &  \textbf{Rewards} & \multicolumn{3}{c}{\textbf{Success rates}} \\
\textbf{Method} &  \textbf{hard} & \textbf{easy} & \textbf{med.} & \textbf{hard} \\
\midrule  %
Modular RL (ours)    & \textbf{11.4} \small{$(0.2)$} & \textbf{90.2\%} & \textbf{78.7\%} & \textbf{78.7\%}  \\
Monolithic RL        & 0.3  \small{$(0.0)$} & 25.0\% & 0.4\%  &  0.0\% \\
Monolithic RL +shared  & \textbf{1.6} \small{$(0.1)$}  & 51.6\% & \textbf{6.8\%} & \textbf{1.5\%}  \\
Monolithic RL +curr  & 0.3 \small{$(0.0)$}  & 56.3\% & 0.0\% & 0.0\% \\
Monolithic RL +shared +curr & 0.8 \small{$(0.0)$} & 69.7\% & 3.9\% & 0.2\% \\
Monolithic RL +easy-only  & 0.6 \small{$(0.1)$} & \textbf{99.0\%} & 1.6\% & 0.4\% \\
\bottomrule
\end{tabular}
}
\tabcaption{\small Comparison results with monolithic RL methods.}
\label{tab:monolithic_comparisons}
\vspace{8pt}
\centering
\scalebox{\tablescaler}{\begin{tabular}{lcccc}
\toprule
 &  \textbf{Rewards} & \multicolumn{3}{c}{\textbf{Success rates}} \\
\textbf{Method} &  \textbf{hard} & \textbf{easy} & \textbf{med.} & \textbf{hard} \\
\midrule
Monolithic RL +top-down                     & 5.1 \small{(0.2)} & 100\% & 51.0\% & \textbf{27.9\%} \\
Structured RL~~+top-down +$\trueabstractstate$  & 11.2 \small{(0.2)} & 100\% & 88.7\%  & \textbf{77.0\%} \\
\bottomrule
\end{tabular}
}
\tabcaption{\small RL results with top-down input.}
\label{tab:topdown_results}
\vspace{8pt}
\scalebox{\tablescaler}{\begin{tabular}{lccc}
\toprule
 & \multicolumn{3}{c}{\textbf{Success rates}} \\
\textbf{Extra inputs} & \textbf{easy} & \textbf{med.} & \textbf{hard} \\
\midrule
top-down                    & 99.8\% & 31.5\%  & \textbf{9.4\%} \\
top-down, $\trueabstractstate$  & 75.2\% & 42.0\%  & \textbf{30.2\%} \\
top-down, $\trueabstractstate$, expert $\instruction$  & 100\% & 78.2\%  & \textbf{54.6\%}  \\
\bottomrule
\end{tabular}
}
\tabcaption{\small Best known RL results from~\cite{mujoban} (SOTA)}
\label{tab:sota}
\end{minipage}%
\hfill%
\begin{minipage}[]{0.34\textwidth}%
\centering
\includegraphics[width=0.69\textwidth]{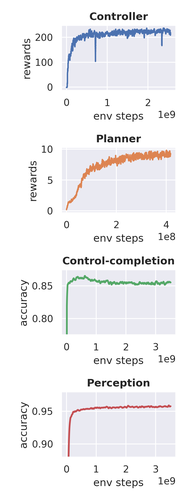}
\captionof{figure}{\small Module learning curves.} %
\label{fig:learning_curves}
\end{minipage}%
\end{table}

To better understand the benefit of prescribed module objectives, future work may compare with general hierarchical RL methods, such as the Option-Critic~\citep{bacon2017option}. 
We found that carefully designing objectives, e.g., for the controller to move to the center of cells had a large impact on overall performance. Learning this from reinforcements alone would likely be an additional significant challenge for general HRL methods.
Further, knowing the abstract state of the puzzle is strictly necessary in many cases for Mujoban, but in early experiments we found that an unstructured LSTM was unable to learn the perception task (even in a supervised setting). This suggests that learning from much more indirect reinforcements, hierarchical RL alone (without task-specific structure or top-down observations) is unlikely to perform well in Mujoban.

\paragraph{Model-based planning and control}
A model-based equivalent of our approach could be considered for Mujoban with classic perception, planning and control modules.  However, even with hard-coded rules and known abstract states, planning for Sokoban (e.g. with MCTS) is known to be computationally expensive~\cite{guez2019investigation}. Further, when the Sokoban state input to the planner is inferred from partial observations it can be (unavoidably) incomplete or unsolvable, for which planning is undefined. Finally, while building a classic controller for simple dynamics, such as for the ball robot, would be certainly possible, it would require significant effort for more complex dynamics, e.g. for the Mujoco ant, especially because of the need to push and align boxes with precision.

\subsection{Module transfer: Mujoban with the Mujoco ant}
\label{sec:ant_results}

Next, we want to validate if RL modules can be similarly reusable as modules of classic robot systems. We consider a scenario where a different robot needs to solve the same task. We replace the default ball robot of Mujoban with the Mujoco ant~(\figref{fig:ant}).  We scaled the physical size of the environment by a factor of three to account for the larger size of the ant, while keeping the weight of boxes and the size of grid pegs unchanged. To focus on the significantly increased control complexity we use the fully-observable Mujoban setting where abstract states are observed, and thus we only need the planner and controller modules. We also increase evaluation time limits to 480s to account for the larger environment and slower motion of the ant.

We train a new controller module for the ant, similarly as for the ball before. We then compose a policy from the ant controller and the old planner module that was trained with the ball robot.  %
Results are in \tabref{tab:ant_results}. Despite the significantly more challenging motor control problem, success rates with the ant are only slightly lower than with the ball body (73.7\% vs. 81.8\%). This suggests that RL modules can be indeed reusable. Such ability to generalize and reuse parts of the solution is an important benefit over monolithic approaches.

\begin{table}[tb]
\begin{minipage}[]{0.4\textwidth}%
\centering
\includegraphics[width=0.9\textwidth]{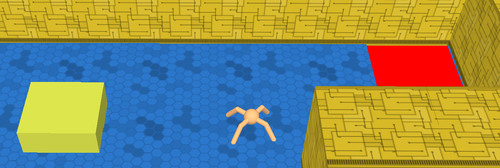}
\captionof{figure}{\small Mujoco ant in Mujoban.}
\label{fig:ant}
\end{minipage}%
\hfill%
\begin{minipage}[]{0.55\textwidth}
\centering
\scalebox{\tablescaler}{\begin{tabular}{lccccc}
\toprule
& \textbf{Robot} & \multicolumn{3}{c}{\textbf{Success rates}} \\
\textbf{Method}  & \textbf{} & \textbf{easy} & \textbf{med.} & \textbf{hard} \\
\midrule
Modular RL & ball (default)  & 100\% & 90.0\% & \textbf{81.8\%} \\
Modular RL & Mujoco ant      & 99.6\% & 83.1\% & \textbf{73.7\%}  \\
\bottomrule
\end{tabular}
}
\tabcaption{\small Module transfer results for the Mujoco ant.}%
\label{tab:ant_results}
\end{minipage}%
\end{table}

\section{Discussion \& Conclusions}

Our results show that coupling RL with an appropriately designed modular learning architecture can lead to progress on domains that are off-puttingly difficult for monolithic approaches. Given enough capacity, computation and experience, a monolithic end-to-end architecture could of course learn to solve Mujoban.
But as our baselines show, for tasks that combine perception, abstract reasoning, and motor control, these costs may be prohibitive and make research progress infeasible.
Modular architectures can facilitate the search for the correct primitives, even in the face of intractable RL tasks. The interpretability of modules allow researchers to make more informed design interventions and incorporate learning signals both more flexibly and more directly.

Modularity is not without its pitfalls: a mis- or over-specified modular structure may lead to unnecessary overhead for learning and may even prevent the agent from finding the optimal solution. 
This pitfall can potentially be avoided by designing the modules to be as general as possible and by allowing joint refinement. 
For embodied tasks, we argue that inherent structural properties can be duly exploited: the world is spatially organized, contains persistent objects, and the agent perceives and acts locally from a single physical location. 
In this spirit, future work should continue to probe the boundary between module engineering and monolithic RL to characterize the best trade-offs between domain-specific engineering and a tabula rasa design.

The modular approach was successful in Mujoban, perhaps because layers of reasoning can be well separated in this domain. Going forward, we want to investigate the same modular philosophy in scenarios with blurrier decompositions, such as assembling furniture~\cite{lee2019ikea}, where independent  training could be followed by joint fine-tuning with respect to the final task objective to compensate for potentially incorrect assumptions. Recent results in this direction are encouraging~\cite{karkus2019differentiable}.

\section{Acknowledgements}

We thank Kenneth Chaney, Bernd Pfrommer, and Kostas Daniilidis for a helpful discussion of the classical robotics literature.

{
\footnotesize
\bibliography{modular_rl}
}

\clearpage
\appendix

\section{Mujoban domain}
\label{sec:app_mujoban}

\myparagraph{Overview.}
Mujoban~\citep{mujoban} is a simulation domain that embeds Sokoban puzzles in Mujoco simulation environment~\cite{todorov2012mujoco}. Sokoban is a popular RL benchmark with complex planning~\cite{junghanns1997sokoban, racaniere2017imagination, guez2019investigation}, where the agent pushes boxes onto target locations in procedurally generated 2D grids.
Mujoban generates 3D maze equivalents of Sokoban levels with varied visual appearance. An embodied agent navigates the maze given partial visual observations, and manipulates boxes using its physical body.

\myparagraph{Simulator.}
The Mujoban domain is visualized in \figref{fig:mujoban}.
In the default configuration the agent has a 2-DoF ball body; and it receives partial observations in the form of first-person camera images $\visual$, standard proprioceptive observations $\proprio$, and the agent's absolute pose $\dense \pose=(x, y, \psi)$.
The simulator also provides access to additional observations, e.g. for training: top-down camera images $\topdown$, absolute pose of boxes $\boxposes$, and abstract state $\abstractstate$ that represents the underlying Sokoban state as a $N\mytimes N\mytimes 4$ binary image, where image channel correspond to the presence of wall, target, agent, and box in each of the grid locations.

\myparagraph{Rewards.}
Rewards in Mujoban are similar to Sokoban: +1 for pushing a box on a target pad, -1 for removing a box from a target pad, and +10 for solving the level.

\myparagraph{Difficulty levels.}
The environment provides different levels of difficulty. In our experiments we aim for the hardest, partially observable configuration of the task:  $10\mytimes10$ Sokoban levels with 4 boxes, first-person observations $(\visual, \proprio, \pose)$. We use easier levels only for evaluation, and leverage extra observations only during training (and for baselines). \tabref{table:mujoban_levels} shows the configuration for each difficulty level. All grids are padded with walls so the final size of all grids are $10\mytimes10$.

\begin{table}[!ht]
    \centering
    \scalebox{0.85}{
    \begin{tabular}{c|c|c|c}
         Difficulty level & Category & Grid size & Number of boxes \\
        \hline
        \hline
        1 & easy & $5\mytimes5$ & 1 \\
        2 & & $7\mytimes7$ & 1 \\
        3 & & $7\mytimes7$ & 2 \\
        4 & medium & $8\mytimes8$ & 3 \\
        5 & hard & $10\mytimes10$ & 4 \\
    \end{tabular}
    }
    \caption{Mujoban difficulty levels}
    \label{table:mujoban_levels}
\end{table}

\myparagraph{Grid pegs.} To make the underlying logic of Mujoban similar to Sokoban, pegs are inserted at each grid point, acting as physical barriers that confine the motion of boxes to horizontal and vertical axes (and make manipulating boxes harder).
Yet the correspondence with Sokoban rules is not perfect, e.g., there is no built-in mechanism that prevents pushing multiple boxes, and although challenging, getting a box off a wall is also conceivable.

\section{Modular RL details}
\label{sec:app_module_details}

\subsection{Perception}
The perception network is a recurrent classifier mapping from $(\visual_{t'}, \pose_{t'})_{t'\leq t}$ and to $\abstractstate_t$.
The perception network is composed of three components: the first one is a neural network with geometry-aware structural prior, shown in \figref{fig:mapping}, which pre-processes each visual frame separately. The output of the network is fed into a ConvLSTM~\cite{xingjian2015convolutional}, which forms an implicit belief state over the abstract state. The output of the LSTM is input into a classifier with 5 labels (None,Wall,Box,Target,Box-on-target), which predicts each spatial location of the abstract state separately. An autoregressive classifier could also be used, but we found it not necessary.

The geometry-aware frame processor works intuitively by attending to different elements of the visual input for each possible location of the abstract state. More precisely, we first extract a $H \mytimes W \mytimes C$ feature tensor from $\visual_t$ using a ResNet. Then, for each spatial location $(x, y)$ of the abstract state $\abstractstate_t$, we use the agent pose input $\pose_t$ to compute the coordinates of the spatial location relative to the pose. The relative spatial coordinates are tiled and concatenated with pixel coordinates to a $H\mytimes W \mytimes C$ features tensor. We pass the resulting coordinate matrix through a CNN, which defines attention weights of size $H \mytimes W \mytimes 1$.
We normalize weights to sum to one, and compute the weighted sum of image features over the $H$ and $W$ dimensions for each channel, which results in a single feature column $1\mytimes1\mytimes C$. This is done for each element $(x,y)$ of the abstract state in parallel (using shared weights); the different outputs are then recombined into a $10\mytimes10\mytimes C$ feature map.

The geometry-aware perception network relies on knowing the agent's pose. We used a setting where the pose is directly observed for simplicity, but it could be also predicted from visual and proprioceptive observations.

We collect data for training by executing the planner, controller and control-completion modules, and using true abstract states as labels. We treat the output of the perception network as independent binary classifiers for each cell, separately for the box, target pad and wall layers of the abstract state. For the agent state layer we treat the  output as a single classifier over all cells. The loss is then given by the sum of independent cross-entropy losses for each classifier.

\subsection{Planner}
The planner is a model-free RL policy for a time-abstracted RL task. Inputs are abstract state estimates, $\estabstractstate_t$, outputs are discrete instructions, $\instruction_t$.

The planner policy is trained with MPO~\cite{abdolmaleki2018maximum}, a recent actor-critic off-policy RL algorithm. 
We adapt MPO to a time-abstracted discrete RL task. In each step of the time-abstracted task the controller module executes the abstract instruction through multiple real environment steps, as many as needed to complete the instruction. Rewards are defined as the sum of real environment rewards collected during the execution of the instruction, plus a $-0.01$ reward for each planner step.
In case the instruction is not completed within $\dense \controltimelimit = 120$ steps, the experience is dropped and treated as an early-termination of the time-abstracted episode. During training we input ground-truth abstract states to the planner, which is replaced by predicted $\estabstractstate_t$ during evaluation.

We use a policy network similar to the repeated ConvLSTM architecture of \citet{guez2019investigation}. The network is built only of generic components, but it has strong structural bias for playing Sokoban.
We use a feed-forward variant of the architecture, replacing the hidden states by a trainable variable that is not propagated through time. We add a separate critic head to the architecture with 5 output values, the Q-value estimates for each discrete instruction. %

\subsection{Controller}
\myparagraph{Control policy.}
The controller is a goal-oriented model-free policy trained with MPO.
Inputs are abstract instruction $\instruction_t$, visual observation $\visual_t$, proprioceptive observations $\proprio_t$, and completion signal $\completion_t$. Outputs are continuous motor torques $\action_t$.

The policy network is a ResNet for processing the visual input connected to an LSTM~\cite{hochreiter1997long} and a fully-connected layer that outputs parameters of a multivariate Normal distribution.
We use an asymmetric actor-critic setup, where the critic network receives privileged information during training. The inputs to the critic are the agent pose $\pose_t$, the pose of boxes $\boxposes_t$, proprioceptive observations $\proprio_t$ and a control action $\action$. The output is a Q-value estimate for $\action$.

During training the controller receives multiple (random) instructions. We define rewards for training the controller as follows.
The agent receives a positive reward ($+4$) when an instruction is completed, and a negative reward ($-5$) if the instruction is not completed within $\dense \controltimelimit=120$ steps. Further, to encourage forward motion a small negative reward is given every step the agent is moving backwards. Rewards are computed using privileged state observations from the simulator.

\myparagraph{Control completion.}
The controller also outputs a completion signal $\completion_t$ using a separate learned classifier. Specifically, the control-completion network is a sequential binary classifier trained with supervised data. The inputs are
$\visual_t, \proprio_t$, $\instruction_{t-1}$ and $\completion_{t-1}$. The output is $p_t(\completion)$, the estimated probability that the controller has completed the last abstract instruction at time $t$. 
The network architecture is a convolutional ResNet connected to an LSTM. In our implementation ResNet weights are shared with the perception network.
The completion-signal $\completion_t$ is sampled according to $\completion_t \sim p_t(\completion)$. During evaluation a positive completion signal is added if there has been no completion predicted for $120$ steps.

We train the control-completion component together with the perception module by executing the planner and controller modules. During execution the predicted $\completion$ signal is fed to the controller. Supervised training labels are obtained according to a strict definition of completing an abstract instruction (defined below), and computed using privileged observations available during training. We use a re-weighted cross-entropy loss that accounts for the  imbalanced number of positive and negative samples. Specifically, we re-weight the loss with weight $w_t = 1 + N_{missed}$, where $N_{missed}$ is the number of steps the prediction has been negative while the label has been positive.

The definition of completing an instruction is as follows. An instruction is completed if the agent is near the center of the target grid cell corresponding to the instructed move, with a $d_{tol}=0.1$ unit tolerance. If the target cell is occupied by a box the box needs to be pushed to the center of the adjacent cell with $d_{tol}=0.1$ tolerance. If the target cell is occupied by a wall the instruction is infeasible, and the agent must remain in its current grid cell. When training the controller we compute the completion criteria and rewards based on privileged state observations. When evaluating the system the completion signal is given by the $\completion_t$ prediction.

\subsection{Training and implementation}\label{sec:app_training}

We train modules of modular RL on randomly generated Mujoban levels from the most difficult category ($10 \mytimes 10$ grid with $4$ boxes). After modules are trained we evaluate the full system on a separate set of 512 random levels. We use privileged observations of the true environment state during training but not during evaluation.  An evaluation episode is successful if the level is solved within 240s (4800 environment steps).
We also run evaluations on easier levels of Mujoban, with smaller grid and less boxes, but we do not use the easier levels during training unless indicated.

Algorithm~\ref{alg:modular_rl} describes a general algorithm for training a modular policy with time abstractions. We omitted the following domain-specific implementation details for clarity. We include side information ($\boxposes$, $\trueabstractstate$) in $\obs$ during training which is used to compute controller rewards and supervised losses, as described under each module.  We use distributed RL where experience collection and module updates are separated into different nodes. While joint training of all modules would be possible, for simplicity we only train one module at a time in a bottom-up sequence (controller, planner, perception). Untrained modules are replaced with ground-truth or random inputs.
The perception module and the completion predictor of the controller are always trained together, and their ResNet weights are shared. Because the perception module is a recurrent network, even if predictions are not needed we have to unroll the network to update its hidden state in the innermost loop Algorithm~\ref{alg:modular_rl}. For terminal states to be meaningful in an abstract episode we drop the last step in case the controller fails to complete an instruction. Finally, we distinguish induced early-termination and environment termination and only do bootstrapping in MPO for the former.

In our distributed RL setup we use one learner and $\numactors$ actors.
We choose $\numactors$ when training each module independently to keep the ratio of actor and learner steps for off-policy learning similar. We use $\numactors = 200$ for the perception module, $\numactors=1000$ for the planner module, and $\numactors=256$ for the controller module.
We train modules until near convergence, which took up to 7 days each. %
Baseline networks were trained in a similar distributed setup, while we continued training for 21 and 5 days, using $\numactors=1000$ and $\numactors=256$, for fully-observable and partially-observably baselines, respectively.  %
For partially observable baselines, following \citet{mujoban}, we terminated episodes early during training ($\dense T_{env}=45s$) which we found to produced better results in early experiments. For evaluation we increase the time limit to $\dense T_{env}=240s$, same as for modular RL.

All models are implemented in Tensorflow~\citep{tensorflow2015-whitepaper} and trained with the Adam optimizer~\citep{kingma2014adam}. We use learner nodes equipped with an Nvidia Tesla V100 GPU. %
Actor nodes have access only to CPU cores.

\section{Baselines}
As an alternative to our modular design, we run MPO with standard monolithic network architectures and train end-to-end for the overall rewards. To gain better understanding of the task difficulty, we also consider a simplified setting with top-down input and train agents with monolithic and structured architectures using the V-Trace~\cite{espeholt2018impala} algorithm. 
While the choice of RL algorithm, input observations, and network architecture are orthogonal, we did not try their combinations due to the substantial computation and time required to run these experiments (on the order of weeks).

\subsection{Baselines for the full, partially-observable Mujoban}
\myparagraph{Monolithic RL.}
We run MPO with standard network architectures that connect a ResNet to an LSTM. 
We search over network variants of three distinct sizes, as well as learning rates, discount factor, MPO parameters, and report the best setting. A small network variant is similar to the controller module except it does not receive instruction inputs. The medium network increases the LSTM hidden state size and the size of fully connected layers. The large network adds extra pre-processing layers for the non-visual inputs and extra layers after the LSTM output. We search over network variants and report best results. %

\myparagraph{Asymmetric vs. symmetric actor-critic.}
For a fair comparison, we first try an asymmetric actor-critic setting, where the critic network receives the same amount of privileged information we used for training modular RL, i.e., the true abstract state $\trueabstractstate_t$, and poses $\pose_t$ and $\boxposes_t$.
The policy network receives the same set of inputs as the modular policy: visual observation, agent pose, and proprioceptive observations. 
The critic network receives the same privileged information as we used for training modular RL, i.e., the true abstract state $\trueabstractstate_t$, and physical state including the absolute pose of the agent and all boxes.
We then also try a symmetric setup, where both actor and critic receive the same input. Here we use a single network torso with separate policy and Q-function network heads.

\myparagraph{Curriculum.}
We train baselines only on hard levels, same as for modular RL, as well as using a training curriculum with levels randomly sampled from all difficulty categories ($p_1=0.25, p_2 = 0.25, p_3 = 0.2, p_4=0.2$,  $p_5=0.1$), and training only on easy levels ($p_1 = 1.0$). After training we evaluate for all difficulty categories.

\subsection{Baselines for simplified, fully-observable Mujoban}

\myparagraph{Fully-observable monolithic RL.}
Here we provide the agent with access to top-down camera images of the full environment, removing (most of) the partial observability. We first use a monolithic LSTM architecture and the V-trace~\cite{espeholt2018impala} algorithm. We train end-to-end for the overall RL task. We use curriculum with combination of all level difficulties.

The network details following \citet{mujoban} are as follows. Proprioception inputs are all concatenated and passed to one layer MLP with 100 hidden units. The vision inputs are passed through 3 layer ResNet with channels sizes 16, 32, 32 and each layer consists of 2 blocks. The outputs flattened and passed through one layer MLP with size of 256. 
All the flat inputs finally concatenated together and passed to and LSTM agent with similar architecture as the controller of our modular RL approach.

\myparagraph{Fully-observable structured RL.}
Finally, we also try an alternative approach with strong structured exploration priors for Mujoban. Here the agent observes both top-down images and the true abstract state.  
We add structured exploration similar to the ``random planner'' agent in~\cite{mujoban}. That is we add an extra pseudo reward of 0.1 for reaching one of the randomly chosen 4 adjacent state in the abstract space in a given time frame. This subgoal is given to the agent by inputting the target abstract state along with the current abstract state.  The agent has extra value head and computes a  V-trace loss separately for the pseudo reward and the sub-task episode. The gradients from this auxiliary loss are added to those associated with the main task, using a weight of 0.5 for the auxiliary loss.
The aim of the auxiliary loss is to help the agent to explore more meaningfully in the abstract  space.  If the subgoal is not reached within 50 steps a new random subgoal is sampled.
The network architecture is similar to the fully-observable monolithic agent. We train using the same curriculum as for monolithic baselines.

\section{Extended literature review}
\label{sec:app_related_work}

\subsubsection*{Deep RL and end-to-end reasoning}

Recent research in deep RL has led to dramatic progress in the components of embodied reasoning, including multi-step abstract reasoning \cite{silver2016mastering,berner2019dota,vinyals2019alphastar}, egocentric perception and state estimation, e.g.,  \cite{levine2016end, merel2019hierarchical,merel2019reusable, gregor2019shaping}, and embodied spatiotemporal control \cite{heess2017emergence,gupta2019relay,andrychowicz2020learning} . In spite of this, and though few in the community deny the importance of the full embodied reasoning problem, little recent work has attempted to tackle problems with all components of embodied intelligence.    Notable exceptions include solving the Rubik's cube using a robotic hand~\cite{andrychowicz2020learning}, and the IKEA RL domain~\cite{lee2019ikea}. To contextualize our aims and approach, we review recent work in RL as it relates to solving full embodied reasoning tasks and contrast it to modular systems design in the wider control literature.

\subsubsection*{Modular systems in AI and robotics}
Although recent research in RL has largely focused on end-to-end solutions to behavioral problems, modular designs have a long history in artificial intelligence research. As early as the 1960s, researchers had designed systems with decomposable structure in order to solve challenging reasoning problems: see (\citealt{Nilsson1980Principles}, \S 1.4) for a discussion. The later robotics literature is full of modular systems that encompass perception and motor control in addition to abstract problem solving, e.g. \citet{Brooks1985Robust, Gat97onthree-layer}. Modular design is one of the cornerstones of practical robotics systems because of the current intractability of end-to-end methods, the need for careful diagnosis of each component, and the desire for component reuse. The systems proposed for DARPA robotics challenge illustrate the general success of modular design patterns: current autonomous and semi-autonomous systems for controlling cars \citep{thrun2006stanley}, humanoid robots \citep{lim2017robot}, and aerial vehicles \citep{Mohta2018Experiments} consist of specialized modules for perception, reasoning, and actuator control coupled together with carefully designed interfaces.

\subsubsection*{Modular structure in deep RL}

Although methods in deep RL often attempt to avoid engineering structure in favor of developing general methods, research into system structure has been a key component of deep RL's success. For example, consider the line of work leading from AlphaGo to MuZero \citep{silver2016mastering, silver2017mastering, silver2018mastering, schrittwieser2019mastering}: the earliest models in this family included domain-specific engineering, such as data augmentation by board rotation and perfect knowledge of game rules. With the lessons learned by the success of earlier methods, later architectures progressively removed and modified structure - removing assumptions about game rules, refining the details of Monte Carlo tree search, and upgrading the convolutional architecture to residual networks - producing more general and more powerful methods. These later innovations were made possible by incorporating lessons learned by first engineering a more restrictively structured method for a very difficult problem of interest.

Compared to a monolithic system designed purely as a black box, a system that exploits modularity can exploit the system designer's intuition about how to structure a problem using specially designed architectures and losses. In this sense, many recently proposed methods have exploited modularity by engineering specialized architecture or losses: for example, by introducing structure exploiting memory \citep{heess2015memory}, map-based reasoning \citep{parisotto2018neural}, prediction  \citep{jaderberg2017reinforcement,gregor2019shaping}, or several such components \citep{wayne2018unsupervised, jaderberg2019human}. This additional structure is typically chosen by appealing to the information content of the environment or correlations with the behavioral task of interest and empirically verified on a difficult benchmark.

Other related work aims to use algorithmic structures as priors for DNN policies \citep{tamar2016value, farquhar2017treeqn, racaniere2017imagination, amos2018differentiable}.
One recent work that combines the benefits of internal system structure with end-to-end learning is the Differentiable Algorithm Network (DAN) of~\citet{karkus2019differentiable}. This model composes differentiable structures for state estimation (vision + filtering), planning, and control and trains them jointly for partially observed map-based navigation tasks. The resulting model can make plans that take into account global map structure while avoiding visually identified obstacles.

Other methods propose to learn a model which is then used to simplify the task of another model in some way.  
For instance, in hierarchical reinforcement learning (HRL), this may correspond to learning motor primitives~\citep{heess2016learning,merel2019reusable}, learning hierarchies of agents which work at different resolution in time and space~\citep{dayan1993feudal,sutton1999between,vezhnevets2017feudal, bacon2018option}, or discovering structured options for exploration~\citep{gregor2016variational}.

In model-based RL, this corresponds to learning a model of the environment, which can then be leveraged by a controller in various ways, such as providing synthetic training data~\citep{ha2018recurrent, kaiser2019model} or simulating trajectories to improve decision making at test time~\citep{hamrick2017metacontrol, hafner2018learning, faust2018prm}, improving learning of policies during training~\citep{silver2017predictron,oh2017value,finn2017deep}, or combining several of these aspects~\citep{schrittwieser2019mastering}.

Yet another family of approach is to learn to predict structured representation of data and used for downstream reasoning and control; this includes methods trained to detect or segment visual structure and use the estimated structure for downstream reasoning~\citep{finn2016deep, kulkarni2019unsupervised, watters2019cobra} and methods that leverage knowledge of object and physics simulators~\citep{Fazelieaav3123,janner2018reasoning,veerapaneni2019entity}.

Finally, a line of research closely related to our proposed approach has sought to partition or structure reasoning, for instance by using an architecture combining reasoning in structured goal spaces with low-level policies ~\citep{nachum2018data,zhang2018composable, alet2018modular}, combining abstract planning with learned symbolic representations and skills~\citep{quack2015simultaneously, konidaris2018skills}, or introducing sub-modules corresponding to a natural task partition~\citep{andreas2016neural}.

\myparagraph{Integrated robot learning domains}

Integrated robotic tasks has been of interest for decades including indoor navigation~\citep{nilsson1984shakey}, autonomous driving~\citep{thrun2006stanley}, performing household tasks~\citep{kaelbling2013integrated, lagriffoul2018platform} and object manipulation with tools~\citep{toussaint2018differentiable}. 
Recent works introduced a number of integrated robot learning domains suitable for large scale RL in simulation~\citep{mujoban, lee2019ikea, savva2019habitat, dosovitskiy2017carla}. 
In this paper we focus on the Mujoban task~\citep{mujoban}. Mujoban exhibits key challenges of embodied reasoning: partial observability, long-horizon reasoning, continuous motor control and object manipulation. %
Compared to other, more realistic domains, an important benefit of Mujoban is the ability to control the difficulty of different layers of embodied reasoning. That is, we can control the complexity of abstract reasoning by changing the size of the underlying Sokoban level and the number of boxes; the complexity of perception by adding top-down or direct state observations; and the complexity of motor-control by replacing the robot body.

\end{document}